\documentclass{article}

\usepackage{microtype}
\usepackage{subcaption}
\usepackage{booktabs}

\usepackage[accepted]{icml2021}
\usepackage{lipsum}
\usepackage{amsthm}
\usepackage{amsfonts}
\usepackage{amsmath}
\usepackage{xspace}
\usepackage{hyperref}
\usepackage{cleveref}
\usepackage{graphicx}
\usepackage{enumerate}
\usepackage{tabularx}
\usepackage{pgf}
\usepackage{pgfplots}
\usepgfplotslibrary{groupplots}
\pgfplotsset{compat=newest}
\usepackage{xcolor}
\usepackage{bbm}

\newtheorem{definition}{Definition}

\newtheorem{theorem}{Theorem}

\usepackage{tikz}
\newcommand*\circled[1]{\tikz[baseline=(char.base)]{
            \node[shape=circle,draw,inner sep=2pt] (char) {#1};}}

\newcommand{\name}[1]{{\textsc{#1}}\xspace}

\newcommand{\MLP}{\name{mlp}}
\newcommand{\SVM}{\name{svm}}

\newcommand{\ICA}{\name{ica}}

\newcommand{\VAE}{\name{vae}}

\newcommand{\FULLVAE}{\name{fullvae}}
\newcommand{\IVAE}{\name{ivae}}
\newcommand{\IDVAE}{\name{idvae}}

\newcommand{\SSIVAE}{\name{ss-ivae}}
\newcommand{\SSIDVAE}{\name{ss-idvae}}
\newcommand{\SSFULLVAE}{\name{ss-fullvae}}

\newcommand{\DSPRITES}{\name{dsprites}}
\newcommand{\CARS}{\name{cars3d}}
\newcommand{\SHAPES}{\name{shapes3d}}
\newcommand{\SMALLNORB}{\name{smallnorb}}

\newcommand{\E}{\mathbb{E}}

\newcommand{\norm}{\mathcal{N}}

\newcommand{\Avect}{\mathbf{A}}
\newcommand{\Bvect}{\mathbf{B}}
\newcommand{\Evect}{\mathbf{E}}

\newcommand{\cvect}{\mathbf{c}}

\newcommand{\fvect}{\mathbf{f}}

\newcommand{\uvect}{\mathbf{u}}

\newcommand{\zvect}{\mathbf{z}}
\newcommand{\xvect}{\mathbf{x}}
\newcommand{\yvect}{\mathbf{y}}

\newcommand{\psivect}{\boldsymbol{\psi}}
\newcommand{\etavect}{\boldsymbol{\eta}}

\newcommand{\Sigmavect}{\boldsymbol{\Sigma}}
\newcommand{\zetavect}{\boldsymbol{\zeta}}
\newcommand{\varepsilonvect}{\boldsymbol{\epsilon}}

\newcommand{\thetavect}{\boldsymbol{\theta}}
\newcommand{\Thetavect}{\boldsymbol{\Theta}}
\newcommand{\phivect}{\boldsymbol{\phi}}

\newcommand{\Tvect}{\mathbf{T}}
\newcommand{\varthetavect}{\boldsymbol{\vartheta}}

\newcommand{\R}{\mathbb{R}}

\icmltitlerunning{An Identifiable Double VAE  For Disentangled Representations}

\begin{document}

\twocolumn[
\icmltitle{An Identifiable Double VAE For Disentangled Representations}

\icmlsetsymbol{equal}{*}

\begin{icmlauthorlist}
\icmlauthor{Graziano Mita}{eur,sap}
\icmlauthor{Maurizio Filippone}{eur}
\icmlauthor{Pietro Michiardi}{eur}
\end{icmlauthorlist}

\icmlaffiliation{eur}{EURECOM, 06410 Biot (France)}
\icmlaffiliation{sap}{SAP Labs France, 06250 Mougins (France)}

\icmlcorrespondingauthor{Graziano Mita}{graziano.mita@eurecom.fr}
\icmlcorrespondingauthor{Maurizio Filippone}{maurizio.filippone@eurecom.fr}
\icmlcorrespondingauthor{Pietro Michiardi}{pietro.michiardi@eurecom.fr}

\icmlkeywords{disentangled representations, machine learning, icml}

\vskip 0.3in
]

\printAffiliationsAndNotice{}  

\begin{abstract}
    A large part of the literature on learning disentangled representations focuses on variational autoencoders (\VAE).
Recent developments demonstrate that disentanglement cannot be obtained in a fully unsupervised setting without
inductive biases on models and data.
However, Khemakhem et al., AISTATS, 2020 suggest that employing a particular form of factorized prior,
conditionally dependent on auxiliary variables complementing input observations, can be one such bias, resulting in
an identifiable model with guarantees on disentanglement.
Working along this line, we propose a novel \VAE-based generative model with theoretical guarantees on identifiability.
We obtain our conditional prior over the latents by learning an optimal representation, which imposes an additional
strength on their regularization.
We also extend our method to semi-supervised settings.
Experimental results indicate superior performance with respect to state-of-the-art approaches, according to several
established metrics proposed in the literature on disentanglement.
\end{abstract}

\section{Introduction}
\label{sec:introduction}
Representation learning aims at learning data representations such that it is easier to extract useful
information when building classifiers or other predictive tasks~\citep{Bengio2013}.
Representation learning seeks to obtain the following properties: i) \textit{expressiveness}: a reasonably-sized
representation should allow to distinguish among a high number of different input configurations;
ii) \textit{abstractness}: learned representations should capture high-level features;
iii) \textit{invariance}: representation should be invariant to local changes of input configurations;
iv) \textit{interpretability}: learned representations should allow each dimension to be informative about the given
task.
These properties are at the core of \textit{disentangled representations}.
\begin{figure}[t!]
    \centering
    \def\svgwidth{\columnwidth}
    \scalebox{0.6}{
\begingroup%
  \makeatletter%
  \providecommand\color[2][]{%
    \errmessage{(Inkscape) Color is used for the text in Inkscape, but the package 'color.sty' is not loaded}%
    \renewcommand\color[2][]{}%
  }%
  \providecommand\transparent[1]{%
    \errmessage{(Inkscape) Transparency is used (non-zero) for the text in Inkscape, but the package 'transparent.sty' is not loaded}%
    \renewcommand\transparent[1]{}%
  }%
  \providecommand\rotatebox[2]{#2}%
  \newcommand*\fsize{\dimexpr\f@size pt\relax}%
  \newcommand*\lineheight[1]{\fontsize{\fsize}{#1\fsize}\selectfont}%
  \ifx\svgwidth\undefined%
    \setlength{\unitlength}{250.64579934bp}%
    \ifx\svgscale\undefined%
      \relax%
    \else%
      \setlength{\unitlength}{\unitlength * \real{\svgscale}}%
    \fi%
  \else%
    \setlength{\unitlength}{\svgwidth}%
  \fi%
  \global\let\svgwidth\undefined%
  \global\let\svgscale\undefined%
  \makeatother%
  \begin{picture}(1,0.52030494)%
    \lineheight{1}%
    \setlength\tabcolsep{0pt}%
    \put(0,0){\includegraphics[width=\unitlength,page=1]{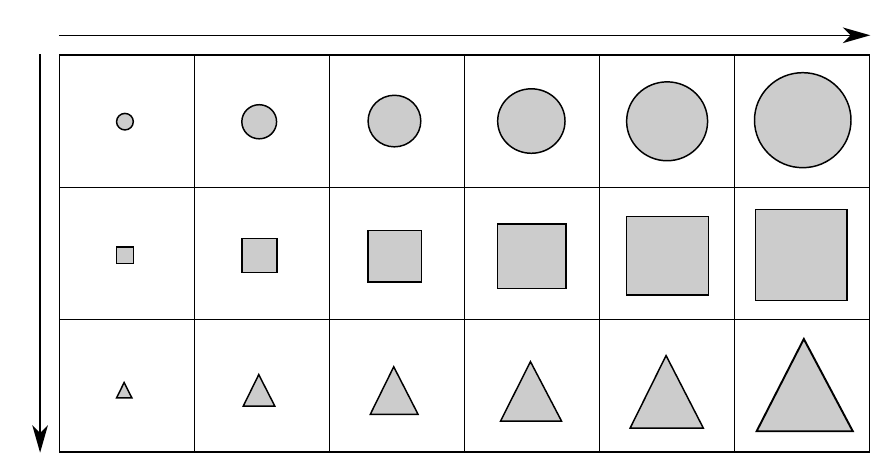}}%
    \put(0.51530371,0.49795642){\color[rgb]{0,0,0}\makebox(0,0)[lt]{\lineheight{2}\smash{\begin{tabular}[t]{l}z1\end{tabular}}}}%
    \put(-0.00189355,0.22330867){\color[rgb]{0,0,0}\makebox(0,0)[lt]{\lineheight{2}\smash{\begin{tabular}[t]{l}z2\end{tabular}}}}%
  \end{picture}%
\endgroup%
}
	\caption{Toy example where each $z_i$ controls a given ground-truth factor: $z_1$ the size, $z_2$ the shape of the 2D objects.}
    \label{fig:toy_disentanglement}
\end{figure}
In disentangled representation learning, the main assumption is that high-dimensional observations $\xvect$ are
the result of a (possibly nonlinear) transformation applied to a low dimensional latent variable of independent
generative factors, called \textit{ground-truth factors}, capturing semantically meaningful concepts.
Input observations can be thought of as the result of a probabilistic generative process, where
latent variables $\zvect$ are first sampled from a prior distribution $p(\zvect)$, and then the observations
$\xvect$ are sampled from $p(\xvect|\zvect)$.
The goal is to learn a representation of the data that captures the generative factors.
In simple terms, illustrated in \cref{fig:toy_disentanglement}, each dimension of a disentangled representation refers
to a single factor of variation.

In this work, we focus on deep generative models, and in particular those based on variational autoencoders (\VAE), to learn disentangled representations.
A well known theoretical result asserts that disentanglement is essentially impossible in a fully unsupervised setting,
without inductive biases on models and data~\citep{Locatello2019}.
However, inducing a disentangled structure into the latent space where $\textbf{z}$ lies is feasible by incorporating auxiliary
information $\textbf{u}$ about the ground-truth factors in the model.
The type and amount of supervision define different families of disentanglement methods, often classified as
supervised, semi-supervised, and weakly-supervised.
In most of these methods, the auxiliary variables $\textbf{u}$ become an integral part of the latent space.
However, recent work~\citep{Hyvarinen2020_IVAE} indicates that there are alternative strategies to
benefit from auxiliary information, such as using it to impose a structure on the latent space.
In their proposal, this is done by learning a prior distribution on the latent space, where the crucial
aspect is that this is conditioned on auxiliary information $\uvect$ that is coupled with every input
observations.
Under mild assumptions, it is possible to show that such form of conditioning implies model
identifiability, allowing one to recover the original ground-truth factors and therefore providing principled disentanglement. 


In this work, we propose a novel generative model that, like~\citet{Hyvarinen2020_IVAE}, uses a conditional prior and
has theoretical identifiability guarantees.
We show that our method naturally imposes an optimality constraint, in information theoretic terms, on the
conditional prior: this improves the regularization on the function that maps input observations to latent variables,
which translates in tangible improvements of disentanglement in practice.
Since assuming to have access to auxiliary variables for each input observations, both at training and testing time, is
not practical in many applications, we also propose a semi-supervised variant of our method.

\noindent{\textbf{Our Contributions}}: i) We present a detailed overview of \VAE-based disentanglement methods using
a unified notation.
Our focus is on the role of the regularization term. 
We introduce a distinction between \textit{direct matching approaches}, in which ground-truth factors are
directly matched to the latent space, and \textit{indirect matching approaches}, where a prior
distribution over the latents is used to structure the learned latent space.
ii) We design a new method, that we call \textit{Identifiable Double \VAE} (\IDVAE) since its ELBO can be seen as a
combination of two variational autoencoders, that is identifiabile, in theory, and that learns an optimal conditional
prior, which is truly desirable in practice.
We additionally propose a semi-supervised version of \IDVAE to make our method applicable also when auxiliary
information is available for a subset of the input observations only.
iii) We design an experimental protocol that uses four well-known datasets, and established disentanglement metrics.
We compare our method to several state-of-the-art competitors and demonstrate that \IDVAE achieves superior
disentanglement performance across most experiments.

\section{Preliminaries}
\label{sec:preliminaries}
Let $\xvect \in \R^n$ be some input observations, which are the result of a transformation of independent latent
ground-truth factors $\zvect \in \R^d$ through a function $\fvect:\R^d \rightarrow \R^n$.
Then, we have that $\xvect = \fvect(\zvect) + \varepsilonvect$, where $\varepsilonvect$ is a Gaussian noise term:
$\varepsilonvect \sim \norm(0,\Sigmavect)$, and independent of $\zvect$.
Let consider the following generative model:
\begin{equation}
    p_{\thetavect}(\xvect,\zvect) = p_{\thetavect}(\xvect|\zvect) p_{\thetavect}(\zvect),
    \label{eq:prob_disentangled_model}
\end{equation}
where $\thetavect \in \boldsymbol{\Theta}$ is a vector of model parameters, $p_{\thetavect}(\zvect) = \prod_{i=1}^d
p_{\thetavect} (z_i)$ represents the factorized prior probability distribution over the latents and $p_{\thetavect}
(\xvect|\zvect)$ is the conditional distribution to recover $\xvect$ from $\zvect$.
The decoder function $\fvect(\zvect)$ determines the way $\zvect$ is transformed into $\xvect$ within $p_{\thetavect}(\xvect|\zvect)$.

Assume to observe some data $\mathcal{D} = \{{\xvect}^{(1)},\cdots,{\xvect}^{(N)}\}$ generated by
$p_{\thetavect^*}(\xvect,\zvect) = p_{\thetavect^*}(\xvect|\zvect) p_{\thetavect^*}(\zvect)$, where 
$\thetavect^*$ are the true, but unknown parameters.
Then, the goal is to learn $\thetavect \in \boldsymbol{\Theta}$ such that:
\begin{equation}
    p_{\thetavect}(\xvect,\zvect) = p_{\thetavect^*}(\xvect,\zvect).
    \label{eq:generative_model}
\end{equation}

When \cref{eq:generative_model} holds, it is then possible to recover the generative ground-truth factors.
Unfortunately, by observing $\xvect$ alone, we can estimate the marginal density $p_{\thetavect}(\xvect) \approx p_{{\thetavect}^*}(\xvect)$, but there are no guarantees about learning the true generative model
$p_{\thetavect^*}(\xvect,\zvect)$.
This is only feasible for models satisfying the following implication:
\begin{equation}
    \forall ({\thetavect},{\thetavect}'): p_{\thetavect}(\xvect) = p_{{\thetavect}'}(\xvect) \Longrightarrow
    {\thetavect} = {\thetavect}'.
    \label{eq:identifiability_condition}
\end{equation}

When \cref{eq:identifiability_condition} holds, the estimated and the true marginal distribution match, and their
parameters match too.
Then, the model is \textbf{identifiable}~\citep{Hyvarinen2020_IVAE} and, as a consequence, it allows one to recover 
the latent ground-truth factors and obtain a disentangled representation:
\begin{equation}\label{eq:identifiability}
    p_{\thetavect}(\xvect) = p_{{\thetavect}'}(\xvect) \Longrightarrow p_{\thetavect}(\xvect,\zvect) =
p_{{\thetavect}'}(\xvect,\zvect).
\end{equation}

A practical goal is to aim for model identifiability \textit{up to trivial transformations}, such as permutation and
scaling; as long as ground-truth factors can be identified, their order and scale is irrelevant.


\section{Related work}
\label{sec:related_work}
Today, a large body of work to learn disentangled representations is based on generative models.
In this work, we focus on \VAE-based approaches~\citep{Kingma2013VAE,Rezende2014VAE}.

\noindent{\textbf{Variational Autoencoder.}} 
A standard \VAE learns the parameters of~\cref{eq:prob_disentangled_model} by introducing an inference model
$q_{\phivect}(\zvect|\xvect)$ to derive an ELBO as follows:
\begin{equation}
    \mathcal{L}_{\VAE} = \E_{q_{\phivect}(\zvect|\xvect)}
    [\log p_{\boldsymbol{\theta}}(\xvect|\zvect)] - \mathrm{KL}(q_{\phivect}(\zvect|\xvect)||p(\zvect)),
    \label{eq:vae_elbo}
\end{equation}
where, by abuse of notation, we write $\xvect$ in place of $\xvect^{(i)}$.
This avoids clutter in the presentation of \VAE-based models, but, clearly, the marginal log-likelihood is
composed of a sum of such ELBO terms, one for each observation $\xvect^{(i)}$~\citep{Kingma2013VAE}.

The distribution $p_{\boldsymbol{\theta}}(\xvect|\zvect)$ has the role of a decoder, whereas $q_{\phivect}
(\zvect|\xvect)$ can be seen as an encoder, and it is generally assumed to be a factorized Gaussian with a diagonal
covariance matrix.
Both distributions are parameterized with neural networks, with parameters $\thetavect$ and variational parameters
$\phivect$.
The prior $p(\zvect)$ is generally a factorized, isotropic unit Gaussian.

The first term of~\cref{eq:vae_elbo} relates to the \textit{reconstruction} of the input data using latent variables
sampled from the variational approximation of the true posterior. 
The second term is a \textit{regularization} term, which pushes the approximate posterior $q_{\phivect}(\zvect|\xvect)
$ to match the prior on the latent space.
Maximizing~\cref{eq:vae_elbo} across observations implies learning the parameters such that the reconstruction
performance is high, and the regularization term is small.

Since both terms that appear in the regularization of \cref{eq:vae_elbo} are factorized Gaussians with diagonal
covariance, one way to interpret the individual components $z_i$ of the latent space is to view them as
independent white noise Gaussian channels~\citep{Burgess2017_AnnealedVAE}.
When the $\mathrm{KL}$ term is zero, the latent channels $z_i$ have zero capacity: this happens when the
approximate posterior $q_{\phivect}(\zvect|\xvect)$ matches exactly the prior $p_{\thetavect}(\zvect)$.
In this case, however, the reconstruction term is penalized.
To increase $\mathrm{KL}(\cdot)>0$, it is necessary to decrease the overlap between channels, and reduce their variances.

\noindent{\textbf{Unsupervised disentanglement learning.}} 
The above understanding of the regularization term is at the basis of many variants of the original \VAE model, that
strive to increase the pressure on the regularization term, or elements thereof, to achieve disentanglement,
without sacrificing reconstruction properties too much.
For example~\citet{Higgins2017_betaVAE} propose $\beta$-\VAE, which modifies~\cref{eq:vae_elbo} by introducing a
hyper-parameter $\beta$ to gauge the pressure on the regularization term throughout the learning process:
\begin{equation}
    \mathcal{L}_{\beta-\VAE} = \mathbb{E}_{q_{\phivect}(\zvect|\xvect)}[\log p_{\boldsymbol{\theta}}(\xvect|\zvect)] - \beta \mathrm{KL}(q_{\phivect}(\zvect|\xvect)||p(\zvect)).
    \label{eq:beta_vae_elbo}
\end{equation}
When $\beta > 1$, the encoder distribution $q_{\phivect}(\zvect|\xvect)$ is pushed towards the unit Gaussian prior $p
(\zvect)$.
In light of the discussion above, the strong penalization of the $\mathrm{KL}$ term in $\beta$-\VAE affects the
latent channel distribution, by reducing the spread of their means, and increasing their variances.

Many methods build on  $\beta$-\VAE~\citep{Burgess2017_AnnealedVAE,Kim2018_FactorVAE,Kumar2018_DIPVAE,
Chen2018_betaTCVAE,Zhao2019_InfoVAE}, rewriting the ELBO in slightly different ways.
A generalization of the $\mathrm{KL}$ term decomposition proposed by~\citet{Hoffman2016_ELBO,Makhzani2017} is the
following~\citep{Chen2018_betaTCVAE}:
\begin{align*}\label{eq:elbo_surgery}
    & \mathbb{E}_{\xvect}[\mathrm{KL}(q_{\phivect}(\zvect|\xvect)||p(\zvect))] =\nonumber \\
    & I(\xvect;\zvect) + \mathrm{KL}(q(\zvect)||\prod_j q(z_j))) + \sum_j \mathrm{KL}(q(z_j)||p(z_j))
\end{align*}
where $q(\zvect)$ is the aggregated posterior and $I(\xvect;\zvect)$ is the mutual information between $\xvect$ and $\zvect$.
Penalizing $I(\xvect;\zvect)$ can be harmful to reconstruction purposes, but enforcing a factorized
aggregated posterior encourages independence across the dimensions of $\zvect$, favouring disentanglement.
The dimensional independence in the latent space is encouraged by the second term, known as total correlation (TC).
The third term is a further regularization, preventing the aggregate posterior to deviate too much from the factorized
prior.

Note that unsupervised \VAE-based approaches approximate the data marginal distribution
$p_{\boldsymbol{\theta}}(\xvect)$, but there are no guarantees to recover the true joint probability distribution
$p_{\boldsymbol{\theta}}(\xvect,\zvect)$, having acces to the input observations $\xvect$
only~\citep{Hyvarinen2020_IVAE}.
Pushing the model to learn a representation with statistically independent dimensions is not a sufficient condition
to obtain full disentanglement.
These considerations were recently formalized in the \textit{impossibility result}~\citep{Locatello2019}, but they
were already known in the nonlinear \ICA literature~\citep{Comon1994ICA,Hyvarinen1999NonlinearICA}.

\noindent{\textbf{Auxiliary variables and disentanglement.}}
To overcome the above limitations, a key idea is to incorporate an inductive bias in the model. The choice of the variational
family and prior distribution can be one of such bias \citep{Mathieu2019,Kumar2020}. Alternatively, it is possible to rely on 
additional information about the ground-truth factors, which we indicate as $\uvect \in \R^{m}$.
When auxiliary observed variables $\textbf{u}$ are available, they can be used jointly with $\zvect$ to reconstruct
the original input $\xvect$.
These methods are usually classified under the semi/weakly supervised family.
More specifically~\citet{Shu2020} identify three 
forms of weak supervision: \textit{restricted
labeling}~\citep{Kingma2014SSVAE,Cheung2015,Siddarth2017SSVAE,Klys2018}, \textit{match/group
pairing}~\citep{Bouchacourt2018_MLVAE,Hosoya2019_GVAE,Locatello2020}, and \textit{rank pairing}~\citep{Chen2020,
Chen2020_ROVAE}.
In the extreme case, when all ground-truth factors are known for all the input samples, we label them as supervised
disentanglement methods.



%

As for unsupervised counterpart, methods relying on auxiliary observed variables $\uvect$ differ in how the
regularization term(s) are designed.
Some approaches use a ``supervised'' regularization term to directly match $\zvect$ 
and the available ground-truth factors $\uvect$: we refer to this form of regularization as \textit{direct
matching}.
An example is what we here call \FULLVAE method~\citep{Locatello2019_SSVAE}, which optimizes the following ELBO:
\begin{equation}\label{eq:full_vae_elbo}
    \mathcal{L}_{\FULLVAE} = \mathcal{L}_{\beta\text{-\VAE}}
	- \gamma R_s(q_{\phivect}(\zvect|\xvect),\uvect),
\end{equation}
where $R_s(\cdot)$ is a loss function between the latent and the ground-truth factors (in the original implementation
it is a binary cross entropy loss). 
Other approaches employ a $\mathrm{KL}$ divergence term between the posterior and the prior over the latents: we
refer to this form of regularization as \textit{indirect matching}.
In other words, direct matching methods require explicit knowledge of one or more ground-truth factors, whereas
indirect matching can also use weak information about them.
\citet{Shu2020} demonstrated that indirect matching methods can enforce some properties in the latent space, leading
to what they define as \textit{consistency} and \textit{restrictiveness}.
To obtain full disentanglement, a method must satisfy both properties on all the latent dimensions.
A recent work by \citet{Hyvarinen2020_IVAE} establishes a theoretical framework to obtain model identifiability, which
is related to disentanglement.
They propose a new generative model called \IVAE, that learns a disentangled representation using a factorized prior
from the exponential family, crucially conditioned on $\textbf{u}$.
In practical applications, the conditional prior is chosen to be a Gaussian location-scale family, where the mean and
variance of each latent dimension $z_i$ are expressed as a function of $\uvect$.
Then, it is possible to derive the following ELBO for the \IVAE model:
\begin{align}
    \mathcal{L}_{\IVAE} &= \mathbb{E}_{q_{\phivect}(\zvect|\xvect,\uvect)}[\log p_{\thetavect}(\xvect|\zvect)]
    \nonumber\\
    &- \beta \mathrm{KL}(q_{\phivect}(\zvect|\xvect,\uvect)||p_{\thetavect}(\zvect|\uvect)).
    \label{eq:ivae_elbo}
\end{align}
In \cref{eq:ivae_elbo}, we recognize the usual structure of a reconstruction, and a regularization term.
A remarkable advancement of the \IVAE model relates to identifiability properties: next, we present a new approach to
learn an identifiable model that leads to disentangled representations, by using an optimal factorized prior,
conditionally dependent on auxiliary observed variables.
We also extend our method to deal with more realistic semi-supervised settings.

\section{\IDVAE: Identifiable Double \VAE}
\label{sec:idvae}

Let $\xvect \in \R^n$, and $\uvect \in \R^m$ be two observed random variables, and 
$\zvect \in \R^d$ a low-dimensional latent variable, with $d \leq n$.
Then, consider the following generative models:
\begin{equation}
	\label{eq:idvae_model}
    p_{\thetavect}(\xvect,\zvect | \uvect) = p_{\fvect}(\xvect|\zvect) p_{\Tvect,\etavect}(\zvect|\uvect),
\end{equation}
\begin{equation}
	\label{eq:idvae_f}
    p_{\fvect}(\xvect|\zvect) = p_{\varepsilonvect}(\xvect - \fvect(\zvect)),
\end{equation}
\begin{equation}
    p_{\Tvect,\etavect}(\zvect|\uvect) = \prod_i h_i(z_i) g_i(\uvect)
    \exp \left[\Tvect_{i}(z_i)^{\top}\etavect_{i}(\uvect) \right],
    \label{eq:conditional_prior}
\end{equation}
and
\begin{equation}
	\label{eq:idvae_prior}
    p_{\varthetavect}(\zvect,\uvect) = p_{\varthetavect}(\uvect|\zvect) p(\zvect),
\end{equation}
where $\thetavect = (\fvect,\Tvect,\etavect)$ and $\varthetavect$ are model parameters.
\Cref{eq:idvae_model} corresponds to the process of generating $\xvect$ given the latents $\zvect$. 
\Cref{eq:idvae_f} implies that $\xvect = \mathbf{f}(\zvect) + \varepsilonvect$, with $\varepsilonvect \sim \norm(0,\Sigmavect)$.
We approximate the injective function $\mathbf{f}$ with a neural network.
\Cref{eq:conditional_prior} is an exponential conditionally factorial distribution~\citep{Bishop2006}, where $h_i$ is
the base measure,
$g_i(\uvect)$ is the normalizing constant, $\Tvect_i = [T_{i,1},\cdots,T_{i,k}]^{\top}$ are the sufficient statistics, and 
$\etavect_i(\uvect) = [\eta_{i,1},\cdots,\eta_{i,k}]^{\top}$ are the corresponding parameters.
The dimension of each sufficient statistic $k$ is fixed.
\Cref{eq:idvae_prior} formalizes the additional process to obtain $\uvect$ given $\zvect$
through $p_{\varthetavect}(\uvect|\zvect)$, where 
$p(\zvect)$ is a prior over the latents, usually a factorized, isotropic unit Gaussian. 

Given a dataset $\mathcal{D} = \{({\xvect}^{(1)},{\uvect}^{(1)}),\cdots,({\xvect}^{(N)},{\uvect}^{(N)})\}$
of observations generated according to \cref{eq:idvae_model,eq:idvae_f,eq:conditional_prior,eq:idvae_prior}, we are interested in finding
a variational bound $\mathcal{L}$ for the marginal data log-likelihood $p(\xvect,\uvect)$, which we derive as follows:
\begin{equation}\label{eq:idvae_derivation1}
	\log p(\xvect,\uvect) = \mathrm{KL}( q_{\phivect}(\zvect|\xvect,\uvect) || p_{\thetavect}(\zvect|\xvect,\uvect)) +
	\mathcal{L}(\thetavect,\phivect), \nonumber
\end{equation}
where, by abuse of notation, we write $\xvect$ and $\uvect$ in place of $\xvect^{(i)}$ and $\uvect^{(i)}$,
which we do hereafter as well.

Since the $\mathrm{KL}$ term is non-negative, we have the following variational lower bound: $\log p(\xvect,\uvect) \geq \mathcal{L}(\thetavect,\phivect)$.
Now, we can write the ELBO, which resembles that of \cref{eq:ivae_elbo}, but includes an additional term:
\begin{align}
	\mathcal{L}(\thetavect,\phivect) &=
    \E_{q_{\phivect}(\zvect|\xvect,\uvect)} [\log p_{\fvect}(\xvect|\zvect)] \nonumber \\
    &- \beta \mathrm{KL}(q_{\phivect}(\zvect|\xvect,\uvect) || p_{\Tvect,\etavect}(\zvect|\uvect)) + \log p(\uvect),
    \label{eq:idvae_derivation3}
\end{align}
where we introduce the parameter $\beta$ to gauge the pressure on the $\mathrm{KL}$ term.
Next, focusing on the generative model in \cref{eq:idvae_prior}, 
we derive the following variational lower bound for $\log p(\uvect)$ in \cref{eq:idvae_derivation3},  
$\log p(\uvect) \geq \mathcal{L}_{\textrm{prior}}(\varthetavect, \psivect)$:
\begin{align}
    \mathcal{L}_{\textrm{prior}}(\varthetavect, \psivect) &=
	\E_{q_{\psivect}(\zvect|\uvect)} [\log p_{\varthetavect} (\uvect|\zvect)] \nonumber \\
	&- \mathrm{KL}(q_{\psivect}(\zvect|\uvect) || p(\zvect)),
	\label{eq:pu_elbo}
\end{align}

Combining \cref{eq:idvae_derivation3} and \cref{eq:pu_elbo},
we obtain:
\begin{align}\label{eq:idvae_full}
	&\mathcal{L}_{\text{\IDVAE}}(\thetavect,\phivect,\varthetavect,\psivect) \geq \nonumber\\
	&\geq \underbrace{\E_{q_{\phivect}(\zvect|\xvect,\uvect)} [\log p_{\fvect}(\xvect|\zvect)] - \beta \mathrm{KL}
	(q_{\phivect}(\zvect|\xvect,\uvect) || p_{\Tvect,\etavect}(\zvect|\uvect))}_{\circled{1}} \nonumber\\
	&+ \underbrace{\E_{q_{\psivect}(\zvect|\uvect)} [\log p_{\varthetavect}(\uvect|\zvect)] -\mathrm{KL}(q_{\psivect}
	(\zvect|\uvect) || p(\zvect))}_{\circled{2}}.
\end{align}
We call our method \IDVAE, Identifiable Double \VAE, because it can be seen as the combination of two variational
autoencoders \circled{1} and \circled{2}, with independent parameters.
In principle, when we optimize the ELBO by summing across all datapoints, e.g. using a doubly stochastic
approach~\citep{titsias14} and automatic differentiation, we could treat the two parts separately.
However, nothing would prevent the conditional prior $p_{\Tvect,\etavect}(\zvect|\uvect)$ and its variational
approximation $q_{\psivect}(\zvect|\uvect)$ to converge to different distributions.
Thus, we further make the modeling assumption of constraining the conditional prior in \circled{1} to be exactly the
variational appoximation learned in \circled{2}, which belongs to the exponential family.

\subsection{Identifiability properties}
\label{sec:idvae_identifiability}

Next, we set up notations and definitions for a general theory of identifiability of generative
models~\citep{Hyvarinen2020_IVAE}, and show that \IDVAE, under mild conditions, is identifiable.

\noindent{\textbf{Notation.}} Concerning the exponential conditionally factorial distribution
in~\cref{eq:conditional_prior}, we denote by $\Tvect(\zvect)$ the vector of concatenated sufficient statistics defined as follows: 
$\Tvect(\zvect) = [\Tvect_1(z_1)^{\top},\cdots,\Tvect_d(z_d)^{\top}]^{\top} \in \R^{dk}$. 
We denote by $\etavect(\uvect)$ the vector of its parameters defined as follows: 
$\etavect(\uvect) = [\etavect_1(\uvect)^{\top},\cdots,\etavect_d(\uvect)^{\top}]^{\top} \in \R^{dk}$.

\begin{definition}
	Let $\sim$ be an equivalence relation on the parameter space $\Thetavect$.
	We say that \cref{eq:prob_disentangled_model} is $\sim$-identifiable if
	$p_{\thetavect}(\xvect) = p_{\thetavect^*}(\xvect) \Longrightarrow \thetavect \sim \thetavect^*$.
	\label{def:identifiability}
\end{definition}
\begin{definition}
    Let $\sim$ be the equivalence relation on $\Thetavect$ defined as follows:
    $(\fvect,\Tvect,\etavect) \sim (\fvect',\Tvect',\etavect') \Leftrightarrow \exists \Avect,\cvect : \Tvect(\fvect^{-1}(\xvect)) = \Avect\Tvect'(\fvect'^{-1}(\xvect)) + \cvect, \forall \xvect \in \mathcal{X}$, where $\Avect$ is a $dk \times dk$ matrix and $\cvect$ is a vector of dimension $dk$. If $A$ is invertible, we denote this relation by $\sim_A$. 
    \label{def:linear_matrix_identifiability}
\end{definition}
\Cref{def:linear_matrix_identifiability} establishes a specific equivalence relation that allows to recover the sufficient statistics of our model up to a linear matrix multiplication. 

\begin{theorem}\citep{Hyvarinen2020_IVAE} 
    Assume we observe data sampled from 
	$p_{\thetavect}(\xvect,\zvect|\uvect) = p_{\fvect}(\xvect|\zvect)p_{\Tvect,\etavect}(\zvect|\uvect)$, where $p_{\fvect}(\xvect|\zvect)$ as in \cref{eq:idvae_f} and $p_{\Tvect,\etavect}(\zvect|\uvect)$ as in \cref{eq:conditional_prior}, 
	with parameters $\thetavect = (\fvect,\Tvect,\etavect)$. Assume the following holds:
    \begin{enumerate}[i]
        \item The set $\{\xvect \in \mathcal{X}:\phi_\epsilon(\xvect)=0\}$ has measure zero, where $\phi_\epsilon$ is the characteristic function of the density $p_\epsilon$ defined in $p_{\fvect}(\xvect|\zvect)=p_{\varepsilonvect}(\xvect - \fvect(\zvect))$.
        \item The function $\fvect$ is injective.
        \item The sufficient statistics $T_{i,j}$ in \cref{eq:conditional_prior} are differentiable almost everywhere, and linearly independent on any subset of $\mathcal{X}$ of measure greater than zero.
        \item Being $k$ the dimensionality of the sufficient statistics $\textbf{T}_i$ in \cref{eq:conditional_prior} and $d$ the dimensionality of $\zvect $, there exist $dk+1$ distinct point $\uvect^0, ..., \uvect^{dk}$ such that the $dk \times dk$ matrix $E$ defined as follows is invertible:
        \begin{equation}
            \Evect=(\etavect(\uvect^1)-\etavect(\uvect^0); \cdots; \etavect(\uvect^{dk})-\etavect(\uvect^0))
            \label{eq:assumption_iv}
        \end{equation}
    \end{enumerate}
    Then the parameters $\thetavect = (\fvect,\Tvect,\etavect)$ are $\sim_A$-identifiable.
    \label{th:identifiability_thorem}
\end{theorem}

\Cref{th:identifiability_thorem} (sketch of the proof in Appendix~C) guarantees a general form of
identifiability for \IDVAE.
Under more restrictive conditions on $\fvect$ and $\Tvect$, following the same reasoning
of \citet{Hyvarinen2020_IVAE}, it is also possible to reduce $\Avect$ to a permutation matrix.

Note that, in practice, all the \VAE-based methods we discuss in this work are approximate.
When using a simple, synthetic dataset, where the generative process is controlled, full disentanglement can be verified
experimentally~\citep{Hyvarinen2020_IVAE}.
However, in a realistic setting, the modeling choice for both $q_{\phivect}(\zvect|\xvect,\uvect)$ and $q_{\psivect}
(\zvect|\uvect)$ can have an impact on disentanglement. Even when recognition models have enough capacity to fit the
data (in our experiments they are Gaussian with diagonal covariance), theoretical guarantees might still fall short,
despite the availability of auxiliary variables for all input observations. This could be due to, for example,
suboptimal solutions found by the optimization algorithm or to the finite data regime.

\subsection{Learning an optimal conditional prior}
\label{sec:idvae_optimal_conditional_prior}

In this paper, we advocate for a particular form of a conditional prior, that is the result of learning an optimal
representation $\zvect$, of auxiliary, observed variables $\uvect$.

In general, an \emph{optimal} representation, for a generic task $\yvect$ (in our case, we aim at reconstructing
$\uvect$) is defined in terms of sufficiency and minimality: $\zvect$ is \textit{sufficient} for the task $\yvect$ if
$I(\uvect;\yvect)=I(\zvect;\yvect)$, where $I(\cdot;\cdot)$ is the mutual information; $\zvect$ is
\textit{minimal} if it compresses the input such that it discards all variability that is not 
relevant for the task \citep{Achille2016}.
As shown in \citep{Tishby1999}, the so called \textit{Information Bottleneck} (IB) can be used to learn an optimal
representation $\zvect$ for the task $\yvect$, which amounts to optimizing the following Lagrangian:
\begin{equation}
	\label{eq:IB}
    \mathcal{L}_{\mathrm{IB}} = H(\yvect|\uvect)+\beta I(\uvect;\zvect),
\end{equation}
where we denote the entropy by $H(\cdot)$, with the constant $\beta$ controlling the trade-off between sufficiency
and minimality.
It is easy to show that~\cref{eq:IB} and~\cref{eq:pu_elbo} are equivalent (with $\beta=1$) when the task is
reconstruction.

In our method, we learn the conditional prior $q_{\psivect}(\zvect|\uvect)$ in part \circled{2} of
\cref{eq:idvae_full}, and use it in part \circled{1} by setting $p_{\Tvect, \etavect} (\zvect|\uvect) = q_{\psivect}
(\zvect|\uvect)$.
In light of above discussion, this is equivalent to imposing an additional constraint that pushes the conditional
prior to learn an \emph{optimal} representation $\zvect$ from $\uvect$; the KL term of part \circled{1} in
\cref{eq:idvae_full} pushes $q_{\phivect}(\zvect|\xvect,\uvect)$ toward the optimal conditional prior, which results
in superior regularization quality.

Note that \Cref{th:identifiability_thorem} requires auxiliary variables $\textbf{u}$ to be expressive enough to recover
all the independent factors through the parameters $\etavect(\uvect)$.
In information theoretic terms, $\uvect$ must be sufficient to recover the ground-truth factors, but there is
no explicit need for the extra optimality constraint on $p_{\Tvect,\etavect}(\zvect|\uvect)$.
While \Cref{th:identifiability_thorem} remains valid for an optimal conditional prior, we demonstrate experimentally
that, when variational approximations, sub-optimal solutions, or finite data size spoil theoretical results, learning
an optimal conditional prior is truly desirable.



\subsection{A semi-supervised variant of \IDVAE}
\label{sec:idvae_semisupervised}
So far, we worked under the assumption that the auxiliary information $\uvect$ is consistently 
available for every $\xvect$. In real scenarios, it is more likely to observe $\uvect$ 
for a subset of the input observations. Thus, we propose a variation of \IDVAE for a semi-supervised setting.
We consider a new objective function that consists of two terms \citep{Kingma2014SSVAE}:
\begin{align}
    \label{eq:ssidvae_elbo} \mathcal{L}_{\SSIDVAE} &= \sum_{(\xvect,\uvect) \sim p_l} \mathcal{L}_{l}(\xvect,\uvect) + \sum_{\xvect \sim p_u} \mathcal{L}_{u}(\xvect),\\
	\label{eq:ssidvae_labeled_elbo} \mathcal{L}_{l}(\xvect,\uvect) &= \mathcal{L}_{\IDVAE}(\xvect,\uvect),\\
    \label{eq:ssidvae_unlabeled_elbo} \mathcal{L}_{u}(\xvect) &= \mathbb{E}_{q_{\zetavect}(\uvect|\xvect)}[\mathcal{L}_{l}(\xvect,\uvect)] + \mathcal{H}(q_{\zetavect}(\uvect|\xvect)), 
\end{align}
where $\mathcal{L}_{l}$ and $\mathcal{L}_{u}$ are the labeled and unlabeled terms respectively;
$q_{\zetavect}(\uvect|\xvect)$ in \cref{eq:ssidvae_unlabeled_elbo} is used to derive 
$\uvect$ from $\xvect$ when $\uvect$ is not provided as input. To be precise, we should add to \cref{eq:ssidvae_elbo} a 
third term -- $\mathbb{E}_{(\xvect,\uvect) \sim p_l}[\log q_{\zetavect}(\uvect|\xvect)]$ -- such that it can learn 
also from labeled data.
Clearly, this method also applies to the work from \citet{Hyvarinen2020_IVAE}.

\section{Experiments}
\label{sec:experiments}
\subsection{Experimental settings}\label{subsec:experimental_settings}

\paragraph{Methods.} We compare \IDVAE against three disentanglement methods: $\beta$-\VAE, \FULLVAE, \IVAE.
$\beta$-\VAE~\citep{Higgins2017_betaVAE} is a baseline for indirect matching methods where no ground-truth factor is
known at training time and the only way to enforce a disentangled representation is by increasing the strength of the
regularization term through the hyper-parameter $\beta$. \FULLVAE~\citep{Locatello2019_SSVAE} is the representative of
direct matching methods: it can be considered as a standard $\beta$-\VAE with an additional regularization term,
weighted by an hyper-parameter $\gamma$, to match the latent space to the target ground-truth factors.
As done in the original implementation, we use a binary cross entropy loss for \FULLVAE, where the targets are normalized
in $[0,1]$; we also set $\beta=1$, to measure the impact of the supervised loss term only. 
\IVAE~\citep{Hyvarinen2020_IVAE} is another indirect matching method where the regularization term, weighted again by $\beta$, involves a 
conditional prior. We additionally report the results for the semi-supervised versions of \FULLVAE, \IVAE, and
\IDVAE, which we denote as \SSFULLVAE, \SSIVAE \footnote{The original work \citep{Hyvarinen2020_IVAE} is not
semi-supervised. We extended it for our comparative analysis.}, \SSIDVAE, respectively.
Variational approximations, and the conditional priors, are Gaussian distributions with diagonal covariance.
All methods have been implemented in PyTorch~\citep{pytorch}.

\paragraph{Datasets.} We consider four common datasets in the disentanglement literature, where observations are
images built as a deterministic function of known generative factors:
\DSPRITES~\citep{Higgins2017_betaVAE}, \SHAPES~\citep{Kim2018_FactorVAE}, \CARS~\citep{Scott2015_cars} and
\SMALLNORB~\citep{LeCun2004_smallnorb}.
We have full control on the generative process and explicit access to the ground-truth factors.
All ground-truth factors are normalized in the range $[0,1]$; for discrete factors, we implicitly assume an ordering
before applying normalization.
All images are reshaped to a 64$\times$64 size.
A short description of the datasets is reported in \cref{tab:datasets}.
Implementations of the generative process for each dataset are based on the code provided by \citet{Locatello2019}.

\paragraph{Disentanglement metrics.} In the literature, several metrics have been proposed to measure
disentanglement, with known advantages and disadvantages, and ability to capture different aspects of disentanglement.
We report the results for some of the most popular metrics: beta score~\citep{Higgins2017_betaVAE},
MIG~\citep{Chen2018_betaTCVAE}, SAP~\citep{Kumar2018_DIPVAE}, modularity and explicitness~\citep{Ridgeway2018}, all
with values between 0 and 1.
The implementation of the metrics is based on \citet{Locatello2019}.
We refer the reader to appendix~E for further details.

\begin{table}[t]
    \scriptsize
    \begin{tabularx}{\columnwidth}{llX}
    \toprule
    \textbf{Dataset}   & \textbf{Size} & \textbf{Ground-truth factors (distinct values)}\\
    \midrule
    \DSPRITES  & 737'280     & shape(3), scale(6), orientation(40), x(32), y(32)\\
    \CARS      & 17'568      & elevation(4), azimuth(24), object type (183)\\
    \SHAPES    & 480'000     & floor color(10), wall color(10), object color(8), object size(8), object type(4), azimuth(15)\\
    \SMALLNORB & 24'300      & category(5), elevation(9), azimuth(18), light(6)\\
    \bottomrule
    \end{tabularx}
    \caption{Main characteristics of the datasets.}
	\label{tab:datasets}
\end{table}

\paragraph{Experimental protocol.} In order to fairly evaluate the impact of the regularization terms, all the tested 
methods have the same convolutional architecture (widely adopted in most recent works), optimizer,
hyper-parameters of the optimizer and batch size.
The latent dimension $\zvect$ is fixed to the true number of ground-truth factors.
The conditional prior in \IVAE is a \MLP network; in \IDVAE we use a simple \MLP \VAE.
The same architecture is taken for the conditional prior of the semi-supervised counterparts.
Moreover, $q_{\zetavect}(\uvect|\xvect)$ is implemented by a convolutional neural network.
We refer the reader to the Appendix~D for more details.

We tried six different values of regularization strength associated to the target regularization term of each method -- $\beta$ for $\beta$-\VAE, \IVAE and \IDVAE, and $\gamma$ for \FULLVAE: $[1, 2, 4, 6, 8, 16]$.
These are recurring values in the disentanglement literature.
For each model configuration and dataset, we run the training procedure with 10 random seeds, given that all methods are susceptible to initialization values.
After 300'000 training iterations, every model is evaluated according to the disentanglement metrics described above.
For \FULLVAE, \IVAE and \IDVAE, all ground-truth factors are provided as input, although \IVAE and \IDVAE work as
well with a subset of them (or with any other additionally observed variable).
We apply the same protocol for the semi-supervised experiments too, where we provide, at training time, 
all the ground-truth factors for a subset of the input observations only, $1\%$ and $10\%$ respectively. At testing time, 
$\uvect$ is instead estimated from $q_{\zetavect}(\uvect|\xvect)$.

\subsection{Experimental results}\label{subsec:experimental_results}

\paragraph{Qualitative Evaluation.} Latent traversal is a simple approach to visualize disentangled representations, by plotting the 
effects that each latent dimension of a randomly selected sample has on the reconstructed output. 
In \cref{fig:dsprites_latent_traversal}, we evaluate a configuration (single seed) of our \IDVAE model trained on \DSPRITES 
(other datasets in Appendix~F). Every row of the figure represents a latent dimension that we vary in the range $[-3,3]$, 
while keeping the other dimensions fixed. We can see that $z_1$ has learned orientation reasonably well; $z_2$ is responsible 
of the object scale; $z_4$ and $z_5$ reflect changes on the vertical and horizontal axis, respectively. $z_3$ 
tried to learn, without success, shape changes. Next, we rely on disentanglement metrics to make a quantitative comparison 
among the tested methods.

\begin{figure}[t!]
    \centering
	\includegraphics[scale=.55]{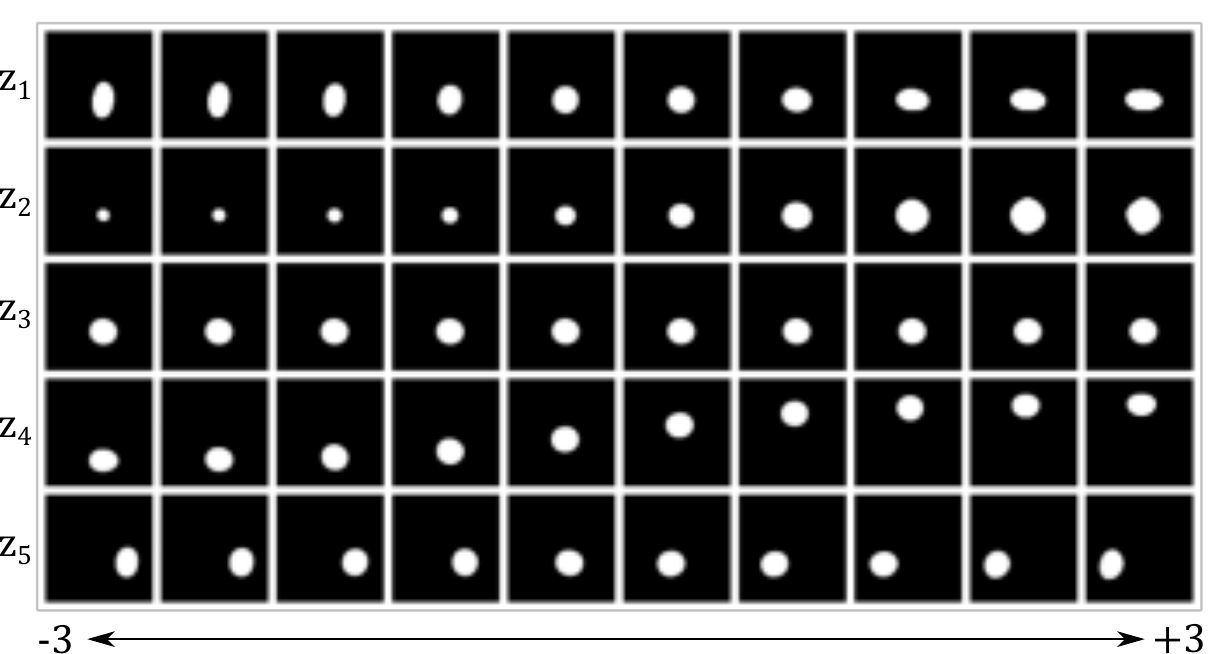}
    \caption{Latent traversal of \IDVAE model trained on \DSPRITES.}
    \label{fig:dsprites_latent_traversal}
\end{figure}

\begin{figure*}[ht]
    \centering
	\begin{subfigure}{\linewidth}
		\resizebox{1\linewidth}{!}{\input{figures/betascores_all.pgf}}
	\end{subfigure}
	\begin{subfigure}{\linewidth}
		\resizebox{1\linewidth}{!}{\input{figures/explicitness_all.pgf}}
	\end{subfigure}
	\caption{Beta score and explicitness (the higher the better). 1=$\beta$-\VAE, 2=\SSIDVAE(1\%), 3=\SSIDVAE(10\%), 4=\IDVAE, 5=\SSIVAE(1\%), 6=\SSIVAE(10\%), 7=\IVAE, 8=\SSFULLVAE(1\%), 9=\SSFULLVAE(10\%), 10=\FULLVAE. Percentage of labeled samples in parenthesis.}
    \label{fig:res1}
\end{figure*}

\begin{figure*}[ht]
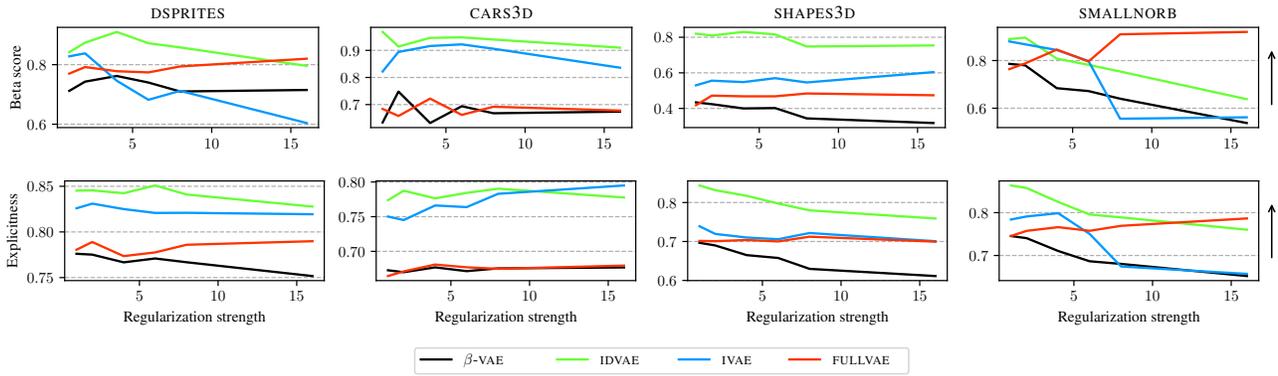

    \centering
	\begin{subfigure}{\linewidth}
		\resizebox{1\linewidth}{!}{\input{figures/betascores_reg_all.pgf}}
    \end{subfigure}
	\begin{subfigure}{\linewidth}
		\resizebox{1\linewidth}{!}{\input{figures/explicitness_reg_all.pgf}}
	\end{subfigure}
	\caption{Beta score and explicitness median (the higher the better) as a function of the regularization strength.}
    \label{fig:res2}
\end{figure*}

\paragraph{Disentanglement Evaluation.} In \cref{fig:res1}, we report, for each method and for each dataset, the
ranges of the beta score and explicitness values with a box-plot.
The variance of the box-plots is due to the random seeds and regularization strengths, which are the only parameters we vary. 
Furthermore, \cref{fig:res1} includes the results for \SSIDVAE, \SSIVAE and \SSFULLVAE (trained with 1\% and 10\% labeled samples), with different shades of green, blue, and red, respectively.
The remaining evaluation metrics can be found in Appendix~F, but they are essentially all
correlated, as also noted in \citet{Locatello2019}.

Overall, we observe, as expected, that $\beta$-\VAE is often the worst method.
Indeed, it has no access to any additional information at training time except the data itself.
Despite this, $\beta$-\VAE disentanglement performance is surprisingly not that far from \FULLVAE that directly
matches the latent space with the ground-truth factors.
In some cases, $\beta$-\VAE obtains very high beta scores (see outliers), such as for \DSPRITES and \CARS datasets,
confirming the sensitivity to random initialization of unsupervised methods~\citep{Locatello2019}.
Note also that \FULLVAE exhibits inconsistent performance across the four datasets.

\IDVAE emerges as the best method across several disentanglement metrics, except for \SMALLNORB, where \FULLVAE's beta score is slightly better.
For this specific dataset and metric, there are no considerable differences among methods, since most of the box-plots overlap.
We note that \IDVAE outperforms \IVAE: considering that the two methods differ for the way the conditional prior is
learned, our experiments show that an optimal conditional prior, as we propose in this work, offers substantial
benefits in terms of disentanglement and it is the only reason for \IDVAE superiority.
Finally, although both \IVAE and \IDVAE have theoretical guarantees on disentanglement and use the
full set of ground-truth factors as input, they do not always obtain the maximum evaluation score, in practice. This is in line with the considerations in \cref{sec:idvae_identifiability}.

The analysis above remains valid if we consider the semi-supervised versions of the tested methods, too. We observe that, with the exception of \SMALLNORB, \SSIDVAE's disentanglement performance  coherently increases when it observes more labeled instances. The same trend is generally followed by \SSIVAE. \SSFULLVAE, instead, seems to be less susceptible to the number of labeled instances. In general, even a small percentage of labeled instances (1\%) is enough for \SSIDVAE to outperform $\beta$-\VAE and to keep up with \FULLVAE that is, however, a fully supervised method. This suggests that \SSIDVAE is a valid choice for applications where collecting additional information about the training data is difficult or expensive.

\paragraph{Impact of the regularization strength.}
The disentanglement performance of each method might change drastically as a function of the regularization strength:
some approaches might work significantly better in some ranges and very badly in others.
In \cref{fig:res2}, we plot, for each method and for each dataset, the median of the beta score and explicitness evaluation
values as a function of the regularization strength.
This is also useful to see if there are methods that consistently dominate others.
In this case, we do not report the results for \SSIDVAE, \SSIVAE, and \SSFULLVAE to make the plots more easily readable.
Additional disentanglement score results, including the semi-supervised versions, can be found in appendix~F.

Across all the datasets, \IDVAE achieves the best median scores for a wide range of regularization strengths.
In \DSPRITES, \CARS, and \SHAPES, \IVAE dominates all the other methods (\IDVAE is largely dominant also considering the remaining evaluations metrics).
The performance of \IVAE and \FULLVAE can match that of \IDVAE in some datasets, but the behavior is not consistent:
if we focus on beta score, \IVAE is the second best method in \CARS and \SHAPES, whereas in \DSPRITES and \SMALLNORB, performance drops when we increase the regularization strength -- even $\beta$-\VAE performs better; \FULLVAE behaves well for \DSPRITES and \SMALLNORB, but it is on pair only with $\beta$-\VAE in \CARS and \SHAPES.

By observing the evolution of the disentanglement scores, it appears that there is no clear strategy to choose the
regularization strength.
For \IDVAE, in datasets such as \DSPRITES and \CARS, the regularization strength does not significantly
affect the beta score; in \SHAPES and \SMALLNORB, we note instead a decreasing monotonic trend. The situation is similar if we look at the explicitness, but it differs if we consider different disentanglement metrics.
It is plausible to deduce that the regularization strength is both model and data specific, and it is also affected by the choice of the disentanglement metric.

\subsection{Limitations}\label{subsec:limitations}
In our experimental campaign we use the same convolutional architecture for all the methods we compare.
We do not vary the optimization hyper-parameters and the dimension of the latent variables.
Hence, we cannot ensure that every method runs in its best conditions.
Nevertheless, our experimental protocol makes our analysis independent of method-specific optimizations, and
has the benefit of reducing training times.

Also, we use the whole set of ground-truth factors as auxiliary variables, in the semi-supervised settings too,
whereas it is possible to study the impact of only a subset of the factors to be available.
Moreover, \IDVAE and \IVAE can use any kind of auxiliary variables, as long as they are informative about the
ground-truth factors: they are not restricted to using, e.g., labels corresponding to input data, as we (and many
other studies) do in our experiments.

Finally, we do not study the implications and benefits of disentanglement for solving complex downstream tasks, which
is an interesting task that we leave for future work.

\section{Conclusion}
\label{sec:conclusion}
In this work, we made a step further in the design of identifiable generative models to learn disentangled
representations. \IDVAE uses a prior that encodes ground-truth factor information 
captured by auxiliary observed variables.
The key idea was to learn an optimal representation of the latent space, defined by an inference network on the
posterior of the latent variables, given the auxiliary variables. Such posterior is then used as a prior on the latent 
variables of a second generative model, whose inference network learns a mapping between input observations and latents.
We also proposed a semi-supervised version of \IDVAE that can be applied when auxiliary variables are available
for a subset of the input observations only. Experimental results offer evidence that 
\IDVAE and \SSIDVAE often outperforms existing alternatives to learn disentangled representations, according to several 
established metrics.

\clearpage{}


\bibliography{references}
\bibliographystyle{icml2021}

\clearpage{}
\onecolumn
\appendix

\section{ELBO derivation for \IDVAE}
\label{sec:idvae_elbo_derivation}
\begin{align}
	\log p(\xvect,\uvect) &= log \int p(\xvect,\uvect,\zvect) d\zvect = \nonumber\\
	&= log \int p(\xvect|\uvect,\zvect) p(\zvect|\uvect) p(\uvect)d\zvect = \nonumber\\
	&= log \int \frac{p(\xvect|\uvect,\zvect) p(\zvect|\uvect) p(\uvect)}{q(\zvect|\xvect,\uvect)}q(\zvect|\xvect,\uvect)d\zvect \geq \mathcal{L}_{\textrm{\IDVAE}}\nonumber\\
	&\geq \mathbb{E}_{q(\zvect|\xvect,\uvect)}[\log \frac{p(\xvect|\uvect,\zvect) p(\zvect|\uvect) p(\uvect)}{q(\zvect|\xvect,\uvect)}] =\nonumber\\
	&= \mathbb{E}_{q(\zvect|\xvect,\uvect)}[\log p(\xvect|\uvect,\zvect)] - KL(q(\zvect|\xvect,\uvect)||p(\zvect|\uvect)) + \log p(\uvect),
	\label{eq:idvae_elbo}
\end{align}
where:
\begin{align}
	\log p(\uvect) &= log \int p(\uvect,\zvect) d\zvect \geq \mathcal{L}_{\textrm{prior}} =\nonumber\\
	&= \mathbb{E}_{q(\zvect|\uvect)}[\log p(\uvect|\zvect)] - KL(q(\zvect|\uvect)||p(\zvect)).
	\label{eq:prior_elbo}
\end{align}
\section{ELBO derivation for \SSIDVAE}
\label{sec:ssidvae_elbo_derivation}
\begin{align}
	\log p(\xvect) &= log \int p(\xvect,\uvect,\zvect) d\uvect d\zvect = \nonumber\\
	&= log \int p(\xvect|\uvect,\zvect) p(\zvect|\uvect) p(\uvect) d\uvect d\zvect = \nonumber\\
	&= log \int \frac{p(\xvect|\uvect,\zvect) p(\zvect|\uvect) p(\uvect)}{q(\uvect,\zvect|\xvect)}q(\uvect,\zvect|\xvect) d\uvect d\zvect \geq\nonumber\\
	&\geq \mathbb{E}_{q(\uvect,\zvect|\xvect)}[\log \frac{p(\xvect|\uvect,\zvect) p(\zvect|\uvect) p(\uvect)}{q(\uvect,\zvect|\xvect)}] =\nonumber\\
	&= \mathbb{E}_{q(\zvect|\xvect,\uvect)q(\uvect|\xvect)}[\log \frac{p(\xvect|\uvect,\zvect) p(\zvect|\uvect) p(\uvect)}{q(\zvect|\xvect,\uvect)q(\uvect|\xvect)}] =\nonumber\\
	&= \mathbb{E}_{q(\uvect|\xvect)}[\mathcal{L}_{\textrm{\IDVAE}}] + \mathcal{H}(q(\uvect|\xvect)).
	\label{eq:ssidvae_unlabeled_component}
\end{align}

Combining \cref{eq:idvae_elbo,eq:prior_elbo,eq:ssidvae_unlabeled_component} we obtain $\mathcal{L}_{\SSIDVAE}$, where it is clear that we use the sum over the data samples instead of the expectation. As stated in the main paper,
we also add the term -- $\mathbb{E}_{(\xvect,\uvect) \sim p_l}[\log q(\uvect|\xvect)]$ -- such that it can learn also from labeled data.
\section{Sketch of the proof of Theorem 1}
\label{sec:theorem_proof}

In this section, we report a sketch of the proof of Theorem 1. Following the proof strategy of \citet{Hyvarinen2020_IVAE}, the proof consists of three main steps.

In the first step, we use assumption (i) to demonstrate that observed data distributions are equal to noiseless 
distributions. Supposing to have two sets of parameters $(\fvect,\Tvect,\etavect)$ and 
$(\tilde{\fvect},\tilde{\Tvect},\tilde{\etavect})$, with a change of variable 
$\bar{\xvect}=\fvect(\zvect)=\tilde{\fvect}(\zvect)$, we show that:

\begin{equation}
	\tilde{p}_{\Tvect,\etavect,\fvect,\uvect}(\xvect) = 
	\tilde{p}_{\tilde{\Tvect},\tilde{\etavect},\tilde{\fvect},\tilde{\uvect}}(\xvect),
\end{equation}
where:
\begin{align}
	\tilde{p}_{\Tvect,\etavect,\fvect,\uvect}(\xvect) &= p_{\Tvect,\etavect}(\fvect^{-1}(\xvect)|\uvect)|det J_{\fvect^{-1}}(\xvect)|\mathbbm{1}_{\mathcal{X}}(\xvect)\label{eq:step_one}
\end{align}

In the second step, we use assumption (iv) to remove all the terms that are a function of $\xvect$ or $\uvect$. 
By substituting $p_{\Tvect,\etavect}$ with its exponential conditionally factorial form, taking the log of both sides of \cref{eq:step_one}, 
we obtain $dk+1$ equations. Then:
\begin{equation}
	\Tvect(\fvect^{-1}(\xvect)) = \Avect\Tvect'(\fvect'^{-1}(\xvect)) + \cvect.
\end{equation}

In the last step, assumptions (i) and (iii) are used to show that the linear transformation is invertible and so 
$(\fvect,\Tvect,\etavect)\sim(\tilde{\fvect},\tilde{\Tvect},\tilde{\etavect})$. This concludes the proof.

For a full derivation of the proof, we point the reader to section $\Bvect$ of the supplement in \citet{Hyvarinen2020_IVAE}, which holds also for our variant of the theorem.
\section{Model architectures, parameters and hyperparameters}
\label{sec:architectures}
All the selected methods (including the semi-supervised variants) share the same convolutional architecture.
The conditional prior in \IVAE is a \MLP network, in \IDVAE we use a simple \MLP \VAE, both with leaky ReLU activation functions.
The ground-truth factor learner implementing $q_{\zetavect}(\uvect|\xvect)$ in \SSIDVAE and \SSIVAE is a convolutional neural network.
\begin{table}[H]
	\centering
    \begin{tabularx}{.85\columnwidth}{ll}
    \toprule
    \textbf{Encoder}   & \textbf{Decoder}\\
    \midrule
	Input: $64\times64\times$ number of channels & Input: $\R^d$, where $d$ is the number of ground-truth factors\\
    $4\times4$ conv, 32 ReLU, stride 2 	& FC, 256 ReLU\\
	$4\times4$ conv, 32 ReLU, stride 2 	& FC, $4\times4\times64$ ReLU\\
	$4\times4$ conv, 64 ReLU, stride 2 	& $4\times4$ upconv, 64 ReLU, stride 2\\
	$4\times4$ conv, 64 ReLU, stride 2 	& $4\times4$ upconv, 32 ReLU, stride 2\\
	FC 256*, FC $2 \times d$ 			& $4\times4$ upconv, 32 ReLU, stride 2\\
										& $4\times4$ upconv, number of channels, stride 2\\
    \bottomrule
    \end{tabularx}
    \caption{Main Encoder-Decoder architecture. In \IVAE and \IDVAE, we give $\uvect$ as input to the fully connected layer of the Encoder which size becomes $256+d$.}
	\label{tab:model_architecture}
\end{table}
\begin{table}[H]
	\centering
    \begin{tabularx}{.6\columnwidth}{ll}
		\toprule
		\textbf{Conditional Prior Encoder}   & \textbf{Conditional Prior Decoder}\\
		\midrule
		FC, 1000 leaky ReLU & FC, 1000 leaky ReLU\\
		FC, 1000 leaky ReLU & FC, 1000 leaky ReLU\\
		FC, 1000 leaky ReLU & FC, 1000 leaky ReLU\\
		FC $2 \times d$ & FC $d$\\
		\bottomrule
    \end{tabularx}
    \caption{\IDVAE Conditional Prior Encoder-Decoder architecture. \IVAE uses the encoder only.}
	\label{tab:conditionalprior_architecture}
\end{table}
\begin{table}[H]
	\centering
    \begin{tabularx}{.75\columnwidth}{l}
		\toprule
		\textbf{Ground-truth Factor Learner}\\
		\midrule
		Input: $64\times64\times$ number of channels. $d$ is the number of ground-truth factors.\\
		$4\times4$ conv, 32 ReLU, stride 2\\
		$4\times4$ conv, 32 ReLU, stride 2\\
		$4\times4$ conv, 64 ReLU, stride 2\\
		$4\times4$ conv, 64 ReLU, stride 2\\
		FC 256, FC $2 \times d$\\
		\bottomrule
    \end{tabularx}
    \caption{Ground-truth factor learner implementing $q_{\zetavect}(\uvect|\xvect)$ in \SSIDVAE and \SSIVAE.}
	\label{tab:groundtruthfactorlearner_architecture}
\end{table}
\begin{table}[H]
	\centering
    \begin{tabularx}{.3\columnwidth}{ll}
    \toprule
    \textbf{Parameters}   & \textbf{Values}\\
    \midrule
	batch\_size & 64\\
	optimizer & Adam\\
	Adam: beta1 & 0.9\\
	Adam: beta2 & 0.999\\
	Adam: epsilon & 1e-8\\
	Adam: learning\_rate & 1e-4\\
	training\_steps & 300'000\\
    \bottomrule
    \end{tabularx}
    \caption{Common hyperparameters to each of the considered methods.}
\end{table}
\section{Implementation of disentanglement metrics}
\label{sec:disentanglement_metrics_params}

\paragraph{Beta score} The idea behind the beta score \citep{Higgins2017_betaVAE} is to fix a random ground-truth factor and sample two mini batches of observations from the corresponding generative model. The encoder is then used to obtain a learned representation from the observations (with a ground-truth factor in common). The dimension-wise absolute difference between the two representation is computed and a simple linear classifier $C$ is used to predict the corresponding ground-truth factor. This is repeated $batch\_size$ times and the accuracy of the predictor is the disentanglement metric score.

\paragraph{MIG - Mutual Information Gap} The mutual information gap (MIG) \citep{Chen2018_betaTCVAE} is computed as the average, normalized difference between the highest and second highest mutual information of each ground-truth factor with the dimensions of the learned representation. As done in \citet{Locatello2019}, we consider the mean representation. and compute the discrete mutual information by binning each dimension of the mean learned representation into $n\_bins$ bins.

\paragraph{Modularity and Explicitness} A representation is modular if each dimension depends on at most one ground-truth factor. \citet{Ridgeway2018} propose to measure the Modularity as the average normalized squared difference of the mutual information of the factor of variations with the highest and second-highest mutual information with a dimension of the learned representation. A representation is explicit if it is easy to predict a factor of variation. To compute the explicitness, they train a one-versus-rest logistic regression classifier to predict the ground-truth factor of variation and measeure its ROC-AUC. In the current implementation, observations are discretized into $n\_bins$ bins.

\paragraph{SAP - Separated Attribute Predictability} According to \citet{Kumar2018_DIPVAE}, the Separated Attribute Predictability (SAP) score is computed from a score matrix where each entry is the linear regression or classification score (in case of discrete factors) of predicting a given ground-truth factors with a given dimension of the learned representation. The (SAP) score is the average difference of the prediction error of the two most predictive learned dimensions for each factor. As done in \citep{Locatello2019}, we use a linear \SVM as classifier.

As explained in the main paper, the implementation of the selected disentanglement evaluation metrics is based on \citet{Locatello2019}. We report the main parameters in \cref{tab:metrics_implementation}.

\begin{table}[H]
    \begin{tabularx}{\columnwidth}{lX}
		\toprule
		\textbf{Disentanglement metrics}   & \textbf{Parameters}\\
		\midrule
		Beta score 	& train\_size=10'000, test\_size=5'000, batch\_size=64, predictor=logistic\_regression\\
		MIG 		& train\_size=10'000, n\_bins=20\\
		Modularity and Explicitness & train\_size=10'000, test\_size=5'000, batch\_size=16, n\_bins=20\\
		SAP score	& train\_size=10'000, test\_size=5'000, batch\_size=16, predictor=linear \SVM, C=0.01\\
		\bottomrule
    \end{tabularx}
    \caption{Disentanglement metrics and their parameters.}
	\label{tab:metrics_implementation}
\end{table}
\clearpage
\section{Full experiments}\label{sec:full_experiments}
In this section, we report the full set of experiments, including reconstructions and latent traversals.
\begin{figure}[H]
    \centering
    \begin{subfigure}{.5\linewidth}
		\centering
        \includegraphics[width=.9\linewidth]{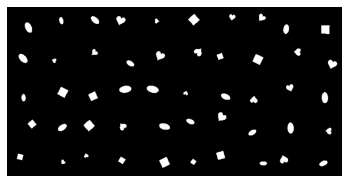}
        \caption{\DSPRITES: original observations.}
    \end{subfigure}%
    \begin{subfigure}{.5\linewidth}
        \centering
		\includegraphics[width=.9\linewidth]{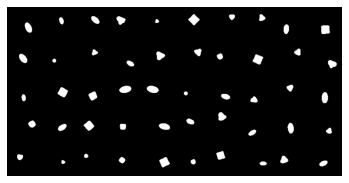}
        \caption{\DSPRITES: reconstructions by \IDVAE.}
    \end{subfigure}
	\begin{subfigure}{.5\linewidth}
		\centering
        \includegraphics[width=.9\linewidth]{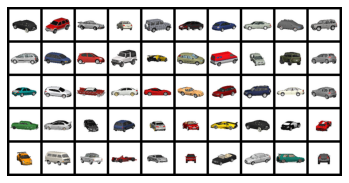}
        \caption{\CARS: original observations.}
    \end{subfigure}%
    \begin{subfigure}{.5\linewidth}
        \centering
		\includegraphics[width=.9\linewidth]{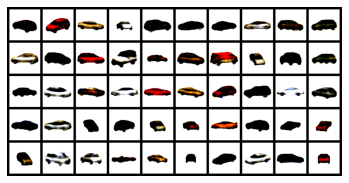}
        \caption{\CARS: reconstructions by \IDVAE.}
    \end{subfigure}
	\begin{subfigure}{.5\linewidth}
		\centering
        \includegraphics[width=.9\linewidth]{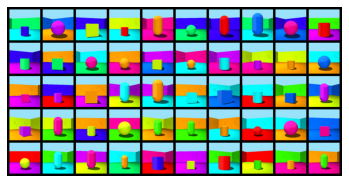}
        \caption{\SHAPES: original observations.}
    \end{subfigure}%
    \begin{subfigure}{.5\linewidth}
        \centering
		\includegraphics[width=.9\linewidth]{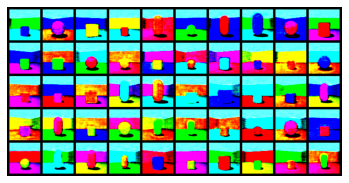}
        \caption{\SHAPES: reconstructions by \IDVAE.}
    \end{subfigure}
	\begin{subfigure}{.5\linewidth}
		\centering
        \includegraphics[width=.9\linewidth]{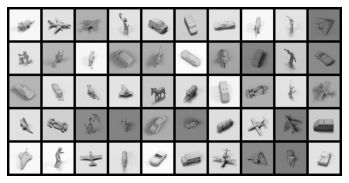}
        \caption{\SMALLNORB: original observations.}
    \end{subfigure}%
    \begin{subfigure}{.5\linewidth}
        \centering
		\includegraphics[width=.9\linewidth]{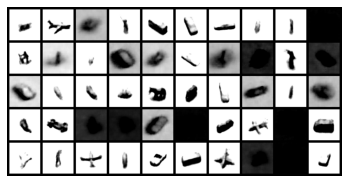}
        \caption{\SMALLNORB: reconstructions by \IDVAE.}
    \end{subfigure}
    \caption{Original observations vs \IDVAE reconstructions.}
\end{figure}

\begin{figure}[H]
    \centering
    \begin{subfigure}[t]{\linewidth}
		\centering
        \includegraphics[scale=.7]{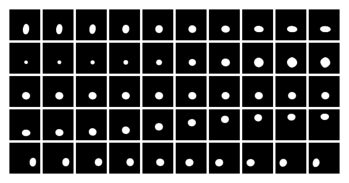}
        \caption{\DSPRITES.}
		\label{fig:dsprites_latent_traversal}
    \end{subfigure}
    \begin{subfigure}[t]{\linewidth}
        \centering
		\includegraphics[scale=.7]{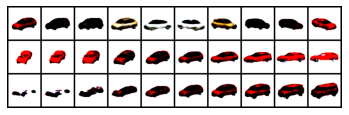}
        \caption{\CARS.}
		\label{fig:cars3d_latent_traversal}
    \end{subfigure}
	\begin{subfigure}[t]{\linewidth}
		\centering
        \includegraphics[scale=.7]{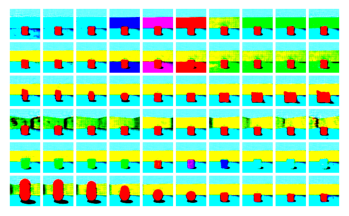}
        \caption{\SHAPES.}
		\label{fig:shapes3d_latent_traversal}
    \end{subfigure}
    \begin{subfigure}[t]{\linewidth}
        \centering
		\includegraphics[scale=.7]{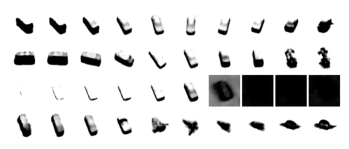}
        \caption{\SMALLNORB.}
		\label{fig:smallnorb_latent_traversal}
    \end{subfigure}
    \caption{\IDVAE latent traversals. Each row corresponds to a dimension of $\zvect$, that we vary in the range $[-3,3]$. We can see that, in some cases, changing a dimension can affect multiple ground-truth factors, meaning that \IDVAE has not obtained full disentanglement. (a) From top to bottom: orientation, scale, shape(?), posY, posX. (b) From top to bottom: azimuth, elevation, object type. (c) From top to bottom: wall color, floor color, object type, azimuth, object color, object size. (d) azimuth, elevation, lighting, category.}
\end{figure}

\begin{figure}[H]
	\begin{subfigure}{\linewidth}
	  \centering
	  \resizebox{\linewidth}{!}{\input{supplement/figures/results/all_betascores.pgf}}
	\end{subfigure}
	\begin{subfigure}{\linewidth}
	  \centering
	  \resizebox{\linewidth}{!}{\input{supplement/figures/results/all_migscores.pgf}}
	\end{subfigure}
	\begin{subfigure}{\linewidth}
	  \centering
	  \resizebox{\linewidth}{!}{\input{supplement/figures/results/all_modularityscores.pgf}}
	\end{subfigure}
	\begin{subfigure}{\linewidth}
	  \centering
	  \resizebox{\linewidth}{!}{\input{supplement/figures/results/all_explicitnessscores.pgf}}
	\end{subfigure}
	\begin{subfigure}{\linewidth}
	  \centering
	  \resizebox{\linewidth}{!}{\input{supplement/figures/results/all_sapscores.pgf}}
	\end{subfigure}
	\caption{Beta score, MIG, Modularity, Explicitness, and SAP (the higher the better). 1=$\beta$-\VAE, 2=\SSIDVAE(1\%), 3=\SSIDVAE(10\%), 4=\IDVAE, 5=\SSIVAE(1\%), 6=\SSIVAE(10\%), 7=\IVAE, 8=\SSFULLVAE(1\%), 9=\SSFULLVAE(10\%), 10=\FULLVAE. Percentage of labeled samples in parenthesis.}
	\label{fig:full_scores}
\end{figure}

\begin{figure}[H]
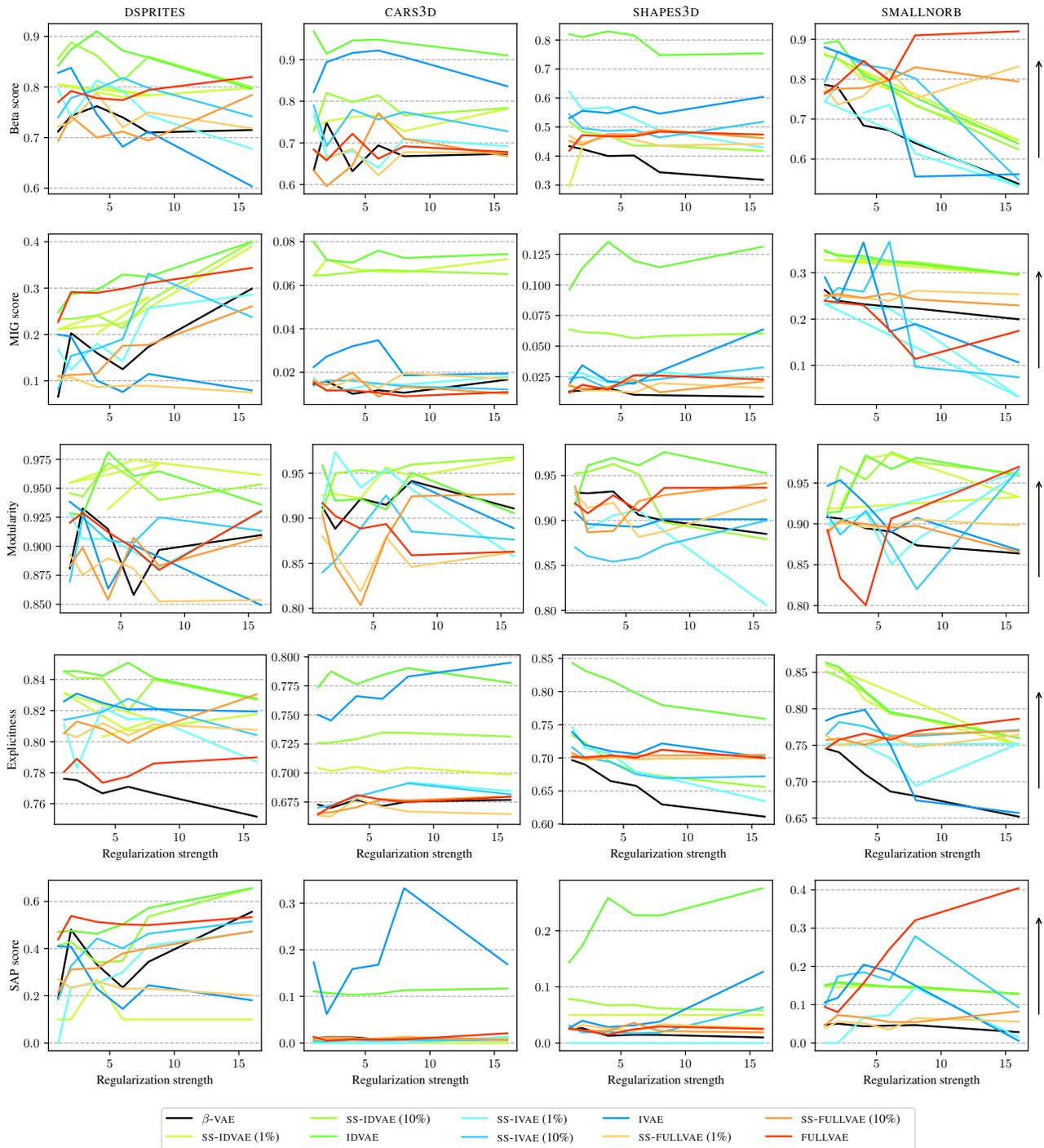

	\begin{subfigure}{\linewidth}
	  \centering
	  \resizebox{\linewidth}{!}{\input{supplement/figures/results/all_betascores_reg.pgf}}
	\end{subfigure}
	\begin{subfigure}{\linewidth}
	  \centering
	  \resizebox{\linewidth}{!}{\input{supplement/figures/results/all_migscores_reg.pgf}}
	\end{subfigure}
	\begin{subfigure}{\linewidth}
	  \centering
	  \resizebox{\linewidth}{!}{\input{supplement/figures/results/all_modularityscores_reg.pgf}}
	\end{subfigure}
	\begin{subfigure}{\linewidth}
	  \centering
	  \resizebox{\linewidth}{!}{\input{supplement/figures/results/all_explicitnessscores_reg.pgf}}
	\end{subfigure}
	\begin{subfigure}{\linewidth}
	  \centering
	  \resizebox{\linewidth}{!}{\input{supplement/figures/results/all_sapscores_reg.pgf}}
	\end{subfigure}
	\caption{Beta score, MIG, modularity, explicitness and SAP median (the higher the better) as a function of the regularization strength, for each method on \DSPRITES, \CARS, \SHAPES, \SMALLNORB.}
	\label{fig:full_scores_reg}
\end{figure}

\end{document}